\documentclass[11pt]{article}

\usepackage[preprint]{acl}

\usepackage{times}
\usepackage{latexsym}
\usepackage[T1]{fontenc}
\usepackage[utf8]{inputenc}
\usepackage{microtype}
\usepackage{inconsolata}
\usepackage{graphicx}
\usepackage{booktabs}
\usepackage{multirow}
\usepackage{xcolor}
\usepackage{amsmath}
\usepackage{amssymb}
\usepackage{xspace}
\usepackage{enumitem}
\usepackage{linguex}
\usepackage{CJKutf8}
\usepackage{pgfplots}
\pgfplotsset{compat=1.18}
\usetikzlibrary{patterns}
\definecolor{decblue}{HTML}{2A78D6}
\definecolor{dmrstorange}{HTML}{EB6834}
\definecolor{iclcviolet}{HTML}{4A3AA7}
\definecolor{iclfaqua}{HTML}{1BAF7A}

\def\repo{\url{https://github.com/larc-iu/dipo}}

\newcommand{\topdown}{\textsc{TopDown}\xspace}
\newcommand{\shiftreduce}{\textsc{ShiftReduce}\xspace}
\newcommand{\dmrst}{\textsc{dmrst}\xspace}
\newcommand{\encdecsr}{\textsc{EncDec-sr}\xspace}
\newcommand{\encdecsexp}{\textsc{EncDec-sexp}\xspace}
\newcommand{\deconlysr}{\textsc{DecOnly-sr}\xspace}
\newcommand{\deconlysexp}{\textsc{DecOnly-sexp}\xspace}
\newcommand{\iclf}{\textsc{icl-f}\xspace}
\newcommand{\iclc}{\textsc{icl-c}\xspace}

\title{Worth Their Weights? Evaluating Generative Models for Discourse Parsing}

\author{Luke Gessler \\
  Indiana University Bloomington \\
  Department of Linguistics \\
  \texttt{lgessler@iu.edu}
\And  
  Amir Zeldes \\
  Georgetown University \\
  Department of Linguistics \\
  \texttt{amir.zeldes@georgetown.edu}}

\begin{document}
\maketitle

\begin{abstract}
Rhetorical Structure Theory (RST) parsing has historically been dominated by discriminative models based on pretrained encoder-only language models.
The recent availability of high-quality generative language models has made it unclear whether RST parsing is better approached as a generative task. 
In this work, we investigate whether generative language models are better at parsing RST than traditional discriminative ones.
We find that at matched parameter counts, discriminative parsers perform much better than generative ones, though much larger generative models sometimes outperform them.
Generative models hold a particular advantage in elementary discourse unit (EDU) segmentation, and we report new state-of-the-art results on the RST-DT benchmark in end-to-end parsing as well as segmentation.
\end{abstract}

\section{Introduction}
\label{sec:intro}

\begin{figure}[t!]
\centering
\includegraphics[width=0.4\textwidth]{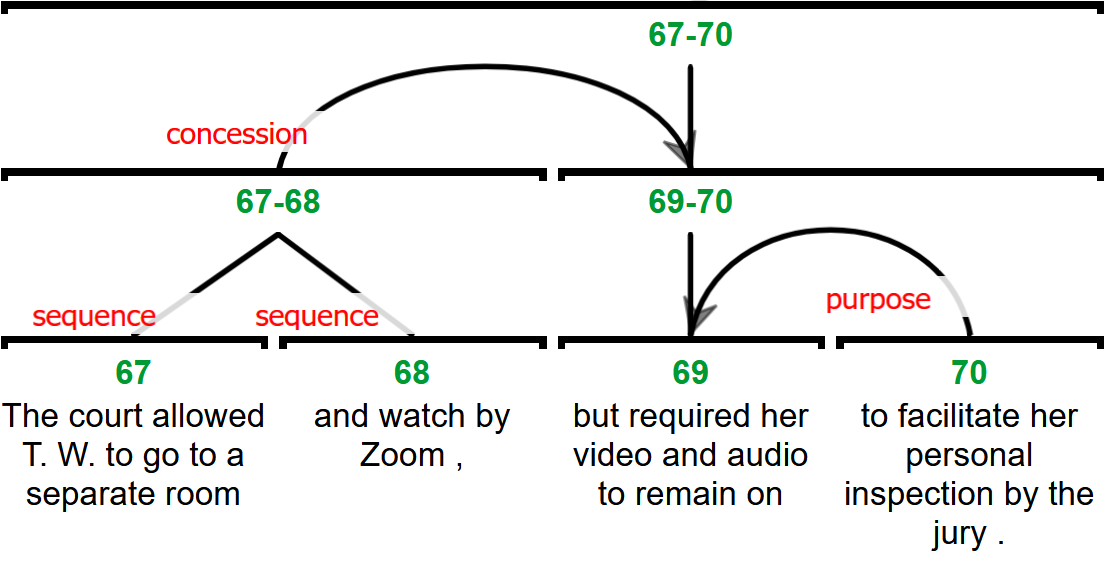}
\vspace{-2pt}
\caption{RST tree fragment from the GUM corpus: The most central EDU $[$69$]$ has a less central \textsc{purpose} satellite, forming unit $[$69-70$]$ which has a \textsc{concession} satellite, itself a temporal \textsc{sequence} of two EDUs.}
\vspace{-6pt}
\label{fig:rst-fragment}
\end{figure}

Natural language texts and conversations are not just homogeneous sequences of sentences: they can be split into meaningful subparts, which play specific roles and may be more or less prominent than one another \cite[1--5]{Stede2012}.
The task of hierarchical discourse parsing consists of segmenting an arbitrary input document into minimal Elementary Discourse Units (EDUs, roughly corresponding to clauses), grouping them into such subparts hierarchically, and labeling their roles using a fixed inventory \cite{marcu-2000}, most commonly in the framework of Rhetorical Structure Theory (RST, \citealt{mann-thompson-1988}, see Figure \ref{fig:rst-fragment}).

Discourse parsing has been used for relation extraction, for example in computational social science \cite{zaczynska-stede-2024-rhetorical} (e.g.\ to identify all \textsc{concession} relations that a politician has conceded in past speeches), as an LLM pretraining/instruction fine-tuning task \cite{iter-etal-2020-pretraining,pu-demberg-2024-rst} or architectural component \cite{bhatia-etal-2015-better,liu-etal-2026-disco,tran-etal-2026-thinkstruct,chen-etal-2026-beyond-chunking}, and for theoretical \cite{Sun2021Investigating,Peng2023CrossParagraph,tsipidi-etal-2024-surprise,Souza2025Taxonomy}, applied \cite{Kong2018Measuring} or literary \cite{RN3358} linguistic research.

RST parsing has historically been dominated by task-specific architectures, which, in the earlier neural age, harnessed pretrained encoders, and usually operated on gold EDU inputs. 
However, in recent years, the model landscape has shifted dramatically.
It is no longer obvious whether the best RST parser is (a) a discriminative parser using a bidirectional encoder, (b) a generative parser obtained by fine-tuning an open-weight LM to emit a linearized tree, or (c) a large LM prompted in context with no training at all.
Further, prior comparisons are confounded by differing data preparation, tree binarization, and evaluation conventions \citep{morey-etal-2017-much}, and most experimental results are difficult to reproduce for researchers, and hard to use for applications due to engineering issues. 

In this paper we investigate 5 research questions:

\begin{enumerate}[itemsep=2pt, parsep=0pt, label=\textbf{RQ\arabic*.}, ref=RQ\arabic*, leftmargin=*]
    \item \label{rq:family} Under matched data and evaluation conditions, which family of parser produces the highest-quality parses in end-to-end and gold EDU settings?
    \item \label{rq:scale} How does performance trade off against the scale of the pretrained language model (PLM) and interact with family?
    \item \label{rq:gen} Within the generative family, how do PLM architecture (encoder-decoder vs. decoder-only) and tree serialization scheme (bottom-up shift-reduce vs. top-down S-expression) each affect parsing quality?
    \item \label{rq:err} What differences in error patterns emerge between families and how much error is attributable to segmentation rather than tree structure and labeling?
    \item \label{rq:multiling} Are the answers to these questions stable across languages and datasets?
\end{enumerate}

To answer these questions, we carry out a comprehensive evaluation controlling for LM and parser architecture, and release a suite of easily usable models for practical RST parsing at state-of-the-art (SotA) or near SotA performance.
Our primary contributions are the following:

\begin{enumerate}[itemsep=2pt, parsep=0pt]
    \item We report rigorous apples-to-apples comparisons of several variants of three model families, showing that encoders outperform alternatives at matched parameter counts and rival much larger decoder models.
    \item We set new SotA results on RST-DT for EDU segmentation ($98.2$ $F_1$) and end-to-end parsing ($55.1$ Full).
    \item We document a previously unnoticed advantage of large generative parsers in EDU segmentation.
    \item We report numbers for five languages using current architectures.
    \item We release code and select off-the-shelf runnable parsers for five languages at \repo{}.
\end{enumerate}

\section{Related Work}
\label{sec:related}

Previous work on neural RST parsing focused on either recursive top-down or bottom up (e.g.~shift-reduce) transition parsing over BERT-style encoders \citep{guz-carenini-2020-coreference,nguyen-etal-2021-rst,kobayashi-etal-2022-simple}, usually evaluating on gold-standard EDU input. 
In the top-down framework, \citet{nguyen-etal-2021-rst} model parsing as a sequence of splitting decisions at token boundaries, so that EDU segmentation falls out of parsing rather than preceding it, and the DMRST parser \citep{liu-etal-2021-dmrst} trains segmentation and top-down parsing jointly. 
Older work has also suffered from inconsistent evaluation paradigms, laid out in detail by \citet{morey-etal-2017-much}, whose evaluation recommendations we adopt.

More recent work using the generative family of models is represented by \citet{hu-wan-2023}, who propose direct text to tree generation for sentence- and document-level parsing with constrained decoding, and \citet{maekawa-etal-2024-obtain}, who fine-tune Llama~2 with QLoRA at up to 70B parameters to predict one parsing action per prompt from the current stack and queue, find bottom-up parsing 1--2 points better than top-down and zero-shot prompting far worse, and report current SotA results on English datasets including the RST Discourse Treebank (RST-DT, \citealt{carlson-etal-2001-building}) and an older version of the Georgetown University Multilayer Corpus (GUM, \citealt{zeldes-2017}).
\citet{li-etal-2024-dialogue} similarly compare plain and transition-based target serializations for LLM-based dialogue discourse parsing.
Multilingual evaluation remains rarer \citep{braud-etal-2017-cross,liu-etal-2021-dmrst,chistova-2025-bridging}, and we are aware of no decoder-based work. 

\section{Data}
\label{sec:data}

As in most previous work, we use two English corpora, RST-DT \citep[LDC2002T07, Wall Street Journal newswire,][]{carlson-etal-2001-building} with the standard 18-relation coarse inventory, under the DISRPT 2025 partition \citep{braud-etal-2025-disrpt} of 309 train, 38 dev, and 38 test documents, and GUM v12.1 \citep[multi-genre written and spoken text,][]{zeldes-2017} with its 32-relation fine-grained inventory, under its official partition (211/32/32, two dev and two test documents from each of sixteen genres).
We evaluate on the full 38-document RST-DT test set.
Following common practice in prior work, trees are binarized right-branching.

To address \ref{rq:multiling} we add one dataset in each of four further languages, using DISRPT 2025 partitions.
For Basque we use the RST Basque TreeBank \citep{iruskieta-etal-2013-rst}, 116/23/24 documents from domains including medical, terminological and scientific.
For German we use the Potsdam Commentary Corpus \citep[PCC,][]{stede-neumann-2014-potsdam}, v2.2 \citep{bourgonje-stede-2020-potsdam}, 141/17/17 newspaper commentaries.
For Persian we use the Persian RST Corpus \citep{shahmohammadi-etal-2021-persian}, 120/15/15 journalistic texts annotated with RST-DT relation classes.
For Chinese we use GCDT \citep{peng-etal-2022-gcdt}, 40/5/5 documents drawn from 5 genres using the same relation inventory as GUM.
These four languages span three scripts and three language families, one of them an isolate.
DISRPT releases segmentation and relation-pair data but not constituent trees, so we take partitions from DISRPT and the trees from each corpus's own release
(see Appendix \ref{app:preproc} for preprocessing details).

\section{Methods}
\label{sec:methods}

\subsection{Parsers}
\label{sec:parsers}

We compare three broad families of parsers: discriminative parsers, fine-tuned generative parsers, and LLMs used as parsers via in-context learning.
All parsers in this work are original implementations, which we release at \repo{}.

\subsubsection{Discriminative parsers}
\label{sec:encoders}

The first family follows the dominant pre-LLM strategy: a pretrained bidirectional encoder yields a contextualized representation of the document's tokens, and task-specific discriminative heads make all structural and labeling decisions.
We compare three variants, all of which use greedy decoding.

\paragraph{\topdown} Following \citet{kobayashi-etal-2022-simple}, we represent each span by the average of its two boundary token representations.
Parsing recursively splits the current span: a deep biaffine scorer over the candidate left and right sub-spans selects the split point, and a second biaffine over the same pair predicts the resulting node's label, which unlike Kobayashi et al.\ we model as a joint nuclearity-and-relation class.

\paragraph{\shiftreduce} The second parser, also based on \citet{kobayashi-etal-2022-simple}, is transition-based, with the same span representation. 
Instead of recursively splitting top-down, a shift-reduce parser is used to successively combine units bottom-up.
The parser is trained with teacher forcing on the gold oracle's action sequence.
Note that \topdown{} and \shiftreduce{} both assume gold EDUs and therefore cannot be applied to raw text.

\paragraph{\dmrst} DMRST \citep{liu-etal-2021-dmrst} is our only discriminative parser capable of ``end-to-end'' parsing, i.e.\ without EDU boundaries as inputs.
Following \citet{chistova-2024-bilingual}, we augment its segmentation module with a linear chain conditional random field.
The parser obtains a mean-pooled representation for each EDU and applies a gated recurrent unit (GRU) encoder to them.
A separate GRU decoder then builds the tree over the EDU representations top-down, as in \topdown, selecting each split by pointer attention \citep{vinyals-etal-2015} over the EDUs inside the current span and labeling it with a biaffine classifier.
Segmentation and parsing train jointly, with the three losses (segmentation, split, label) combined by dynamic weighting.

\paragraph{PLMs}

Pretrained representation quality is essential for parser performance \citep{kobayashi-etal-2022-simple}.
We use a recent English model, Ettin \citep{weller2025seqvsseqopen}, at 150M, 400M, and 1B parameters.
We choose Ettin for its high quality and because it also comes with decoder-only variants of each model trained identically, which allows us to cleanly control for architecture (cf.\ Table~\ref{tab:arch}).
We train only \dmrst{}, the strongest discriminative parser, on all 3 model sizes, and train \topdown{} and \shiftreduce{} on only Ettin-400M in order to keep experiment count manageable.
For the four non-English languages we use XLM-R base \citep[270M parameters,][]{conneau-etal-2020-unsupervised}, since Ettin is English-only.
We discuss our model choices and give full settings in Appendix~\ref{app:disc}.

\subsubsection{Generative parsers}
\label{sec:generative}
\begin{figure}[t]
\centering
\small
\begin{tabular}{@{}p{0.96\columnwidth}@{}}
\toprule
\textbf{RST tree:}
EDUs: $e_1{=}$\emph{The cat sat.}, $e_2{=}$\emph{It was tired.}, $e_3{=}$\emph{So it slept.}

Structure: $\big((e_1\ e_2)_{\textsc{sn}:\text{elaboration}}\ \ e_3\big)_{\textsc{ns}:\text{background}}$.\\
\midrule
\emph{Fine-tuned generative parsers}\\
\addlinespace[2pt]
\textbf{Shift-reduce (\textsc{sr}):}\newline
\texttt{\footnotesize The cat sat . <shift> It was tired . <shift> \textbf{<reduce\_sn\_elaboration>} So it slept . <shift> \textbf{<reduce\_ns\_background>}}\\
\addlinespace[3pt]
\textbf{S-expression (\textsc{sexp}):}\newline
\texttt{\footnotesize \{ \{ \{ The cat sat . \} \{ It was tired . \} \textbf{SN elaboration <label\_end>} \} \{ So it slept . \} \textbf{NS background <label\_end>} \}}\\
\midrule
\emph{In-context learning}\\
\addlinespace[2pt]
\textbf{Segmentation output:}\newline
\texttt{\footnotesize The cat sat . \textbf{||} It was tired . \textbf{||} So it slept .}\\
\addlinespace[3pt]
\textbf{Parsing output:}\newline
\texttt{\footnotesize The cat sat . <shift> It was tired . <shift> \textbf{SN elaboration <label\_end>} So it slept . <shift> \textbf{NS background <label\_end>}}\\
\bottomrule
\end{tabular}
\caption{A three-EDU tree and its serializations.
For the fine-tuned generative parsers, every element in \texttt{< >} is a single special token added to the PLM's tokenizer.
ICL takes two calls: the model marks EDU boundaries in the raw text, then parses its own segmented output. For ICL, \texttt{< >} encloses ordinary strings.}
\label{fig:serial}
\label{fig:icl-sr}
\end{figure}

RST parsing can also be cast as sequence generation: the model reads a document's raw text and generates a serialization of its tree.
\citet{hu-wan-2023} do this with an S-expression serialization, but their decoding constraints still admit unbalanced brackets, which have to be repaired in postprocessing.
We devise two novel serialization schemes, shown in Figure~\ref{fig:serial} and detailed in Appendix~\ref{app:gen}, which overcome this limitation.

\textsc{sexp} nests EDUs in curly braces, with each node's nuclearity and relation spelled out as a postfix in normal tokens after its children.
\textsc{sr} encodes the tree as the actions of a shift-reduce parser, with each \texttt{reduce} and its nuclearity--relation label fused into a single new token.
In both, the source text is reproduced by a special \texttt{copy} action that emits the next source token, rather than by constraining the vocabulary as Hu and Wan do.
Constrained decoding guarantees that every output is well formed, and also lets us evaluate generative parsers with gold EDUs by forcing an EDU boundary exactly at each gold boundary.
All generative parsers are trained with a curriculum that grows from small subtrees to full documents.

\paragraph{PLMs}
We use T5Gemma~2 1B-1B \citep{t5gemma2-2025}, Gemma~3 1B \citep{gemma3-2025}, Gemma~4 31B \citep{gemma4-2026}, and Qwen~3.6 27B \citep{qwen36-2026}, training each model on both serialization schemes.

\subsubsection{In-context learning}
\label{sec:icl}

The third family frames RST parsing as an in-context learning (ICL) task, where the model is given $k$ gold-annotated input--output pairs from the training split of the given corpus and a full label inventory before it makes its predictions.

We exclusively use a variant of the shift-reduce serialization scheme, as shown in Figure~\ref{fig:icl-sr}.
Preliminary experiments showed that models tend to struggle with performing segmentation and parsing simultaneously, and we therefore use a two-step approach where the model first segments and then is prompted again with its own output to perform parsing with the predicted EDUs.
We formulate two variants of this approach below.

\paragraph{\iclf{} (\textbf{F}rontier)} A large near-frontier LLM is used through a hosted API, with no constraint on decoding.
No commercial provider offers grammar-constrained decoding expressive enough for our serialization schema, so the model must adhere to it unassisted.
Any malformed output sequence is recorded as a failure and is treated as a wholly incorrect prediction.
We use DeepSeek V4 Pro \citep[$1.6$T parameters,][]{deepseekv4-2026}, served in \texttt{fp8} through a hosted API accessed via OpenRouter.
We prefer an open-weight model over an API-only model such as those provided by OpenAI and Anthropic in the interest of reproducibility.

\paragraph{\iclc{} (\textbf{C}onstrained)} A self-hosted open-weight LLM is used via llama.cpp \citep{llamacpp}.
The models we self-host are relatively small, but llama.cpp offers a sophisticated grammar-constrained decoding feature, which allows us to enforce our serialization schema so that malformed outputs are impossible.
We use Qwen3.5-397B-A17B and Qwen3.5-122B-A10B \citep{qwen35-2026}.
Due to hardware constraints, we use quantized versions of both: UD-Q4\_K\_XL and Q8\_0, respectively.
See Appendix~\ref{app:icl} for details.

\subsection{Training and inference}
\label{sec:training}

Every discriminative and generative parser is adapted with LoRA \citep{hu-etal-2022-lora} (rank 16, $\alpha=32$), which a preliminary comparison on RST-DT finds comparable to full fine-tuning.
The discriminative parsers are optimized with AdamW \citep{loshchilov2019decoupled} and the generative ones with Adafactor \citep{shazeer-stern-2018}.
Every parser decodes greedily, as is standard in RST parsing---beam search helps only configurations already well below the best (Appendix~\ref{app:beam}).
Hyperparameters follow prevailing practice, with tuning confined to RST-DT development data and a single PLM per family.
Due to the large number of experiments in this work, each configuration is trained and evaluated only once.
See Appendices~\ref{app:disc} and~\ref{app:gen} for full details.

\subsection{Evaluation}
\label{sec:eval}

We report micro-averaged Parseval F-scores (S, N, R, F) over the $n-1$ internal constituents of each binarized $n$-EDU tree: span, nuclearity, relation, and the Parseval full F-score, which we refer to below simply as \emph{Full}.
Following \citet{morey-etal-2017-much}, we exclude trivial single-EDU leaf spans from the counts and key each constituent on its parent span rather than scoring one entry per child, as in RST-Parseval.
For end-to-end systems we additionally report segmentation quality (Seg) as per-token BeginSeg $F_1$, computed with the official scorer of the DISRPT 2025 shared task \citep{braud-etal-2025-disrpt}.
A predicted segmentation that does not reproduce the document text exactly is projected onto the gold token stream, and a document with no prediction scores zero while staying in the denominator.

\section{Results}
\label{sec:results}

\begin{table*}[p]
\centering
\footnotesize
\setlength{\tabcolsep}{4pt}
\begin{tabular}{llcccccccccc}
\toprule
& & \multicolumn{5}{c}{\textbf{RST-DT} (coarse)} & \multicolumn{5}{c}{\textbf{GUM} (fine)} \\
\cmidrule(lr){3-7} \cmidrule(lr){8-12}
\textbf{System} & \textbf{PLM} & Seg & S & N & R & F & Seg & S & N & R & F \\
\midrule
\multicolumn{12}{l}{\emph{Discriminative (LoRA fine-tuned encoder)}} \\
\dmrst & Ettin-150M & 96.0 & 72.5 & 60.9 & 49.9 & 48.6 & 93.7 & 66.6 & 52.8 & 40.5 & 40.3 \\
\dmrst & Ettin-400M & 96.7 & 75.2 & 64.6 & 54.3 & 52.5 & 94.7 & 72.3 & 58.4 & 46.7 & 46.5 \\
\dmrst & Ettin-1B & 96.5 & 75.7 & 64.5 & 53.5 & 52.0 & 94.8 & 71.1 & 58.0 & 46.5 & 46.3 \\
\midrule
\multicolumn{12}{l}{\emph{Generative (LoRA fine-tuned LM)}} \\
\encdecsr & T5Gemma 2 1B-1B & 97.8 & 72.2 & 62.4 & 52.6 & 51.4 & 95.4 & 65.1 & 50.5 & 39.4 & 39.2 \\
\encdecsexp & T5Gemma 2 1B-1B & 95.1 & 66.0 & 56.6 & 47.2 & 46.0 & 90.5 & 53.6 & 43.3 & 33.9 & 33.7 \\
\deconlysr & Gemma 3 1B & 96.7 & 65.9 & 55.7 & 46.6 & 45.7 & 93.9 & 55.9 & 40.9 & 31.3 & 31.1 \\
\deconlysexp & Gemma 3 1B & 92.5 & 50.4 & 42.0 & 35.6 & 34.4 & 76.1 & 30.6 & 23.3 & 18.1 & 17.9 \\
\deconlysr & Gemma 4 31B & 98.2 & 76.7 & 66.4 & 56.4 & 55.1 & 95.7 & 68.2 & 54.3 & 44.0 & 43.9 \\
\deconlysr & Qwen 3.6 27B & 98.2 & 76.3 & 65.9 & 56.3 & 54.9 & 95.9 & 71.2 & 57.7 & 46.7 & 46.4 \\
\deconlysexp & Gemma 4 31B & 94.9 & 66.5 & 57.4 & 48.8 & 47.5 & 90.6 & 55.6 & 45.1 & 36.7 & 36.6 \\
\deconlysexp & Qwen 3.6 27B & 95.8 & 65.4 & 57.5 & 49.1 & 47.5 & 92.0 & 59.4 & 48.6 & 38.8 & 38.6 \\
\midrule
\multicolumn{12}{l}{\emph{In-context (no training)}} \\
\iclf{} (unconstrained) & DeepSeek V4 Pro & 66.3 & 46.3 & 36.2 & 30.3 & 28.2 & 60.2 & 40.4 & 28.4 & 19.8 & 19.4 \\
\iclc{} (grammar) & Qwen 3.5 397B & 89.4 & 54.4 & 33.5 & 30.2 & 24.0 & 84.7 & 48.4 & 27.1 & 17.5 & 15.3 \\
\iclc{} (grammar) & Qwen 3.5 122B & 86.5 & 40.7 & 24.7 & 20.2 & 17.5 & 83.1 & 36.8 & 19.7 & 11.1 & 10.0 \\
\bottomrule
\end{tabular}
\caption{End-to-end parsing results on RST-DT and GUM.}
\label{tab:e2e}
\end{table*}

\begin{table*}[p]
\centering
\footnotesize
\begin{tabular}{llcccccccc}
\toprule
& & \multicolumn{4}{c}{\textbf{RST-DT} (coarse)} & \multicolumn{4}{c}{\textbf{GUM} (fine)} \\
\cmidrule(lr){3-6} \cmidrule(lr){7-10}
\textbf{System} & \textbf{PLM} & S & N & R & F & S & N & R & F \\
\midrule
\multicolumn{10}{l}{\emph{Discriminative (LoRA fine-tuned encoder)}} \\
\topdown & Ettin-400M & 78.3 & 67.4 & 56.2 & 54.8 & 76.9 & 63.1 & 50.3 & 50.0 \\
\shiftreduce & Ettin-400M & 77.0 & 66.5 & 54.7 & 53.5 & 76.0 & 62.7 & 50.7 & 50.3 \\
\dmrst & Ettin-150M & 76.3 & 65.2 & 54.2 & 52.6 & 72.5 & 59.5 & 47.1 & 46.8 \\
\dmrst & Ettin-400M & 78.0 & 68.1 & 57.3 & 55.5 & 76.4 & 63.7 & 51.5 & 51.3 \\
\dmrst & Ettin-1B & 79.5 & 69.1 & 58.1 & 56.5 & 75.7 & 63.6 & 51.6 & 51.4 \\
\midrule
\multicolumn{10}{l}{\emph{Generative (LoRA fine-tuned LM)}} \\
\encdecsr & T5Gemma 2 1B-1B & 74.7 & 64.8 & 55.1 & 53.9 & 68.8 & 54.5 & 42.4 & 42.3 \\
\encdecsexp & T5Gemma 2 1B-1B & 64.8 & 55.5 & 46.7 & 45.5 & 55.0 & 45.6 & 36.1 & 36.0 \\
\deconlysr & Gemma 3 1B & 69.5 & 59.1 & 49.7 & 48.7 & 60.9 & 46.3 & 35.0 & 34.9 \\
\deconlysexp & Gemma 3 1B & 49.4 & 42.1 & 35.4 & 34.5 & 36.5 & 29.0 & 22.6 & 22.5 \\
\deconlysr & Gemma 4 31B & 77.8 & 68.3 & 58.5 & 57.2 & 72.0 & 59.4 & 48.1 & 48.0 \\
\deconlysr & Qwen 3.6 27B & 78.3 & 68.2 & 58.6 & 57.3 & 74.4 & 61.7 & 50.0 & 49.7 \\
\deconlysexp & Gemma 4 31B & 66.3 & 57.8 & 49.7 & 48.3 & 54.3 & 45.0 & 36.5 & 36.3 \\
\deconlysexp & Qwen 3.6 27B & 65.3 & 57.2 & 49.7 & 48.1 & 58.4 & 48.5 & 39.0 & 38.9 \\
\midrule
\multicolumn{10}{l}{\emph{In-context (no training)}} \\
\iclf{} (unconstrained) & DeepSeek V4 Pro & 41.5 & 33.2 & 27.3 & 25.2 & 35.7 & 26.6 & 18.9 & 18.3 \\
\iclc{} (grammar) & Qwen 3.5 397B & 60.9 & 39.3 & 33.6 & 28.1 & 56.4 & 32.7 & 23.2 & 20.1 \\
\iclc{} (grammar) & Qwen 3.5 122B & 51.3 & 31.5 & 27.1 & 21.7 & 48.8 & 28.2 & 16.0 & 14.6 \\
\bottomrule
\end{tabular}
\caption{Gold EDU parsing results on RST-DT and GUM.}
\label{tab:goldedu}
\end{table*}

\begin{table*}[p]
\centering
\footnotesize
\begin{tabular}{l ccc ccc ccc ccc}
\toprule
 & \multicolumn{6}{c}{RST-DT} & \multicolumn{6}{c}{GUM} \\
\cmidrule(lr){2-7}\cmidrule(lr){8-13}
 & \multicolumn{3}{c}{\dmrst{} (enc.)} & \multicolumn{3}{c}{\deconlysr{} (dec.)} & \multicolumn{3}{c}{\dmrst{} (enc.)} & \multicolumn{3}{c}{\deconlysr{} (dec.)} \\
\cmidrule(lr){2-4}\cmidrule(lr){5-7}\cmidrule(lr){8-10}\cmidrule(lr){11-13}
Ettin & Seg & E2E & Gold & Seg & E2E & Gold & Seg & E2E & Gold & Seg & E2E & Gold \\
\midrule
150M & 96.0 & 48.6 & 52.6 & 92.6 & 31.5 & 38.3 & 93.7 & 40.3 & 46.8 & 89.6 & 19.2 & 26.1 \\
400M & 96.7 & 52.5 & 55.5 & 94.5 & 35.3 & 41.3 & 94.7 & 46.5 & 51.3 & 91.9 & 22.5 & 27.6 \\
1B   & 96.5 & 52.0 & 56.5 & 95.7 & 39.6 & 44.4 & 94.8 & 46.3 & 51.4 & 93.0 & 27.9 & 32.5 \\
\bottomrule
\end{tabular}
\caption{Parsing results using matched Ettin encoders and decoders: segmentation $F_1$ (Seg), and Full end to end (E2E) and with gold EDUs (Gold).}
\label{tab:arch}
\end{table*}

\begin{table*}[t]
\centering
\footnotesize
\setlength{\tabcolsep}{5pt}
\begin{tabular}{llccccccccc}
\toprule
& & \multicolumn{5}{c}{\textbf{End-to-end}} & \multicolumn{4}{c}{\textbf{Gold EDU}} \\
\cmidrule(lr){3-7} \cmidrule(lr){8-11}
\textbf{System} & \textbf{PLM} & Seg & S & N & R & F & S & N & R & F \\
\midrule
\multicolumn{11}{l}{\emph{Basque (RST Basque TreeBank)}} \\
\dmrst & XLM-R base & 88.1 & 62.5 & 42.3 & 28.2 & 27.4 & 71.1 & 50.1 & 33.4 & 33.1 \\
\encdecsr & T5Gemma 2 1B-1B & 89.4 & 52.5 & 32.8 & 22.2 & 21.9 & 58.1 & 34.6 & 24.3 & 24.0 \\
\deconlysr & Qwen 3.6 27B & 92.7 & 58.8 & 31.8 & 19.2 & 18.6 & 61.3 & 33.4 & 19.4 & 19.1 \\
\midrule
\multicolumn{11}{l}{\emph{German (PCC)}} \\
\dmrst & XLM-R base & 91.1 & 55.9 & 30.2 & 19.0 & 18.2 & 64.2 & 36.5 & 23.8 & 23.1 \\
\encdecsr & T5Gemma 2 1B-1B & 95.8 & 57.5 & 29.9 & 15.5 & 15.1 & 62.3 & 35.0 & 18.8 & 18.5 \\
\deconlysr & Qwen 3.6 27B & 95.9 & 66.8 & 27.7 & 16.6 & 16.6 & 71.9 & 29.2 & 18.5 & 18.1 \\
\midrule
\multicolumn{11}{l}{\emph{Persian (Persian RST Corpus)}} \\
\dmrst & XLM-R base & 91.6 & 59.6 & 43.3 & 34.4 & 33.1 & 63.8 & 48.5 & 38.4 & 37.1 \\
\encdecsr & T5Gemma 2 1B-1B & 94.0 & 56.0 & 38.2 & 32.1 & 29.9 & 59.2 & 43.5 & 35.1 & 33.6 \\
\deconlysr & Qwen 3.6 27B & 94.2 & 62.6 & 44.2 & 35.8 & 34.9 & 65.8 & 48.2 & 40.5 & 39.4 \\
\midrule
\multicolumn{11}{l}{\emph{Chinese (GCDT, fine)}} \\
\dmrst & XLM-R base & 87.2 & 60.6 & 42.7 & 32.0 & 31.8 & 73.2 & 55.6 & 44.9 & 44.9 \\
\encdecsr & T5Gemma 2 1B-1B & 89.2 & 46.3 & 29.2 & 21.6 & 21.3 & 58.9 & 39.3 & 30.6 & 30.5 \\
\deconlysr & Qwen 3.6 27B & 91.4 & 49.5 & 31.9 & 22.4 & 22.4 & 57.0 & 39.7 & 30.8 & 30.8 \\
\bottomrule
\end{tabular}
\caption{Parsing results on non-English languages.}
\label{tab:multiling}
\end{table*}

For English, Table~\ref{tab:e2e} reports end-to-end results, and Table~\ref{tab:goldedu} reports gold EDU results.
All \iclc{} and \iclf{} runs do poorly, despite their great size.
\iclf{} fails to produce a well-formed parse for 9 of 38 RST-DT documents and for 14 (end to end) and 15 (gold EDU) of 32 GUM documents, which score zero, while the grammar lets \iclc{} parse every document.
On RST-DT, the 27--31B \deconlysr{} parsers set new states of the art in both EDU segmentation ($98.2$ $F_1$) and end-to-end parsing ($55.1$ Full), as detailed below.

\paragraph{Segmentation}
Our RST-DT setting corresponds to the plain track of the DISRPT segmentation task, which supplies tokenized documents without sentence boundaries.
Our largest \deconlysr{} models reach $98.2$, which is to the best of our knowledge a new state of the art: previously, the records without and with gold sentence boundaries available were $97.9$ \citep{chistova-2024-bilingual} and $97.62$ \citep{braud-etal-2023-disrpt}, respectively.
GUM Seg values are not directly comparable with published scores, since DISRPT 2025 used an earlier GUM release.

Predicted EDUs cost \dmrst{} $3.0$--$4.5$ points on the Full metric (F) on RST-DT and $4.8$--$6.5$ on GUM, but the \textsc{sr} parsers on T5Gemma, Gemma, and Qwen only lose $2.1$--$3.0$ and $3.1$--$4.1$ (\ref{rq:err}), so part of the generative family's advantage in structure in end-to-end settings comes from a reduction in error propagation from segmentation.
The generative models' edge in segmentation lies mostly at sentence-internal boundaries without preceding punctuation, where the 27--31B \textsc{sr} parsers recall $97.0$--$97.9\%$ of RST-DT boundaries---compare with \dmrst{} at 400M and 1B $92.8$--$93.4\%$ (Table~\ref{tab:seg-context}).
See Appendix~\ref{app:analysis} for further analysis.

\paragraph{Structure}
With gold EDUs, the best discriminative and generative parsers are roughly tied: the 27--31B generative parsers lead \dmrst{} by under a point on RST-DT, and \dmrst{} leads by $1.7$ on GUM, with 27 times fewer parameters.
Only generative parsers benefit from scale (\ref{rq:scale}): \dmrst{} gains little beyond 400M, while generative parsers improve sharply from 1B to 27--31B.
Wide constituents are responsible for much of the poor performance of smaller models: all parsers correctly identify $91$--$94\%$ of two-EDU constituents on RST-DT, but at 9--16 EDUs \dmrst{} with Ettin-1B identifies $58.7\%$, the 27--31B parsers $50.5$--$53.8\%$, and the Gemma~3 and T5Gemma parsers $32.7$--$42.3\%$ (Table~\ref{tab:width}).

Within the generative family (\ref{rq:gen}), \textsc{sr} beats \textsc{sexp} in every pairing, and encoder--decoder beats decoder-only in the 1B--2B range.
On RST-DT, the 27--31B \deconlysr{} parsers set a new end-to-end state of the art, at $55.1$ Full, surpassing the $53.5$ of \citet{chistova-2025-bridging}, whose model trains jointly on 18 treebanks in 11 languages.
With gold EDUs, the best published result remains the $58.1$ of \citet{maekawa-etal-2024-obtain}'s expensive prompt-per-action 70B Llama~2, which \dmrst{} with Ettin-1B approaches within $1.6$ at two orders of magnitude fewer parameters, while exceeding the best prior encoder-based parser \citep[$55.7$,][]{chistova-2024-bilingual}.

\paragraph{Matched recipe encoder vs. decoder}
\label{sec:arch}

Table \ref{tab:arch} shows the results for the matched Ettin encoder/decoder PLMs trained with the best parser architecture in each family (\dmrst{} and \deconlysr{}).
At matched parameter counts and pretraining, the encoder-based parser outperforms the decoder-based one by a wide margin, in line with the finding of \citet{weller2025seqvsseqopen} that encoders yield better representations for classification tasks (\ref{rq:family}).

\paragraph{Non-English languages}
\label{sec:multiling}

In Table~\ref{tab:multiling} (cf.\ Appendix~\ref{app:multiling-prior} for prior work), \dmrst{} outperforms both generative parsers in three of the four languages, both end to end and with gold EDUs. 
The exception is Persian, where the 27B decoder leads by about two points with 100 times the parameters of XLM-R base.
This is a clearer discriminative win than in English, where the families were roughly tied, which may reflect how much smaller these treebanks are than RST-DT or GUM.
Segmentation again shows the opposite pattern (\ref{rq:err}): both generative parsers segment better than \dmrst{} in every language, so \dmrst{}'s lead comes despite worse segmentation.
Both English patterns therefore hold across languages: \dmrst{} is the more parameter-efficient parser, and the generative parsers segment better (\ref{rq:multiling}).

\section{Error Analysis}
\label{sec:analysis}

\begin{figure}[!tb]
\centering
\includegraphics[trim={14cm 0 14cm 0},clip,width=\columnwidth]{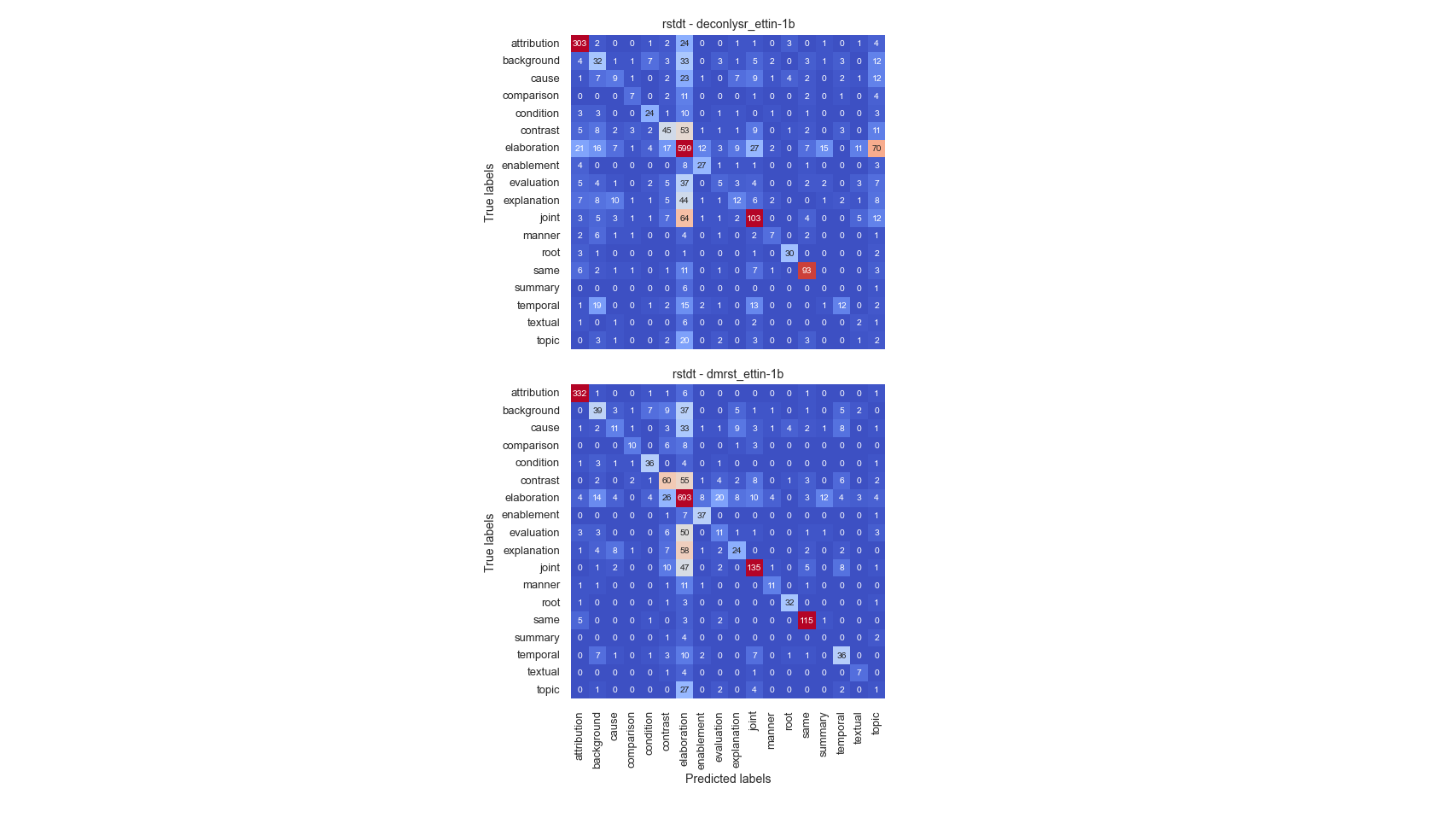}
\caption{Label confusion matrices for top Ettin decoder and encoder models on RST-DT.}
\label{fig:rstdt-ettin-confmats}
\end{figure}

\paragraph{Relation labels} Figure \ref{fig:rstdt-ettin-confmats} shows the confusion matrix of relation labels in RST-DT for the two Ettin models (decoder and encoder), based on a dependency conversion of the data following \citet{li-etal-2014-text}, which looks for each EDU at whether its outgoing discourse relation is correct, regardless of hierarchical structure. 
This allows us to compare which discourse relation semantics get confused for which by each model in a local context. Figure \ref{fig:rstdt-icl-confmats} in the appendix shows the same information for the two best (but overall much worse) ICL approaches. 

The figure shows that while both models overpredict majority classes, especially \textsc{elaboration}, the encoder is much better at avoiding false predictions of \textsc{joint} (a simple additive `A and B' relationship). 
The decoder also has a notable hot spot of 70 false positive \textsc{topic} predictions for true \textsc{elaboration}. \textsc{topic} includes question--answer pairs, rhetorical questions, problem-solution pairs and responses to statements. 
Both types of relations typically occur without explicit discourse markers (DMs) such as `but' or `although', which make \textsc{comparison} relations easier. 
It is unclear whether decoder errors are simply due to mapping of lexical items as in example \ref{ex:topic}, where a trigger word such as `dealing', which might occur in problem-solution relations, could nudge the model to choose \textsc{topic}.

\ex. \textit{We are dealing with real entrepreneurs} $\xleftarrow[\text{decoder:topic}]{\text{gold+DMRST:elab.}}$\textit{There are a lot of them} (wsj\_1146)\label{ex:topic}

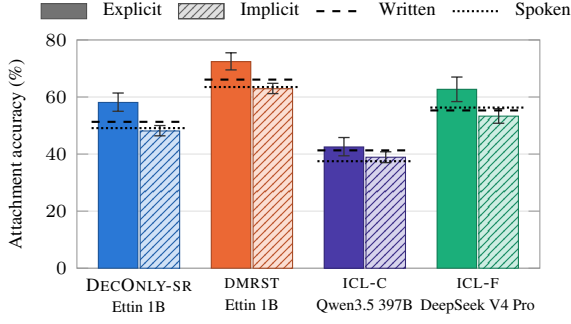
\begin{figure}[t]
\centering
\begin{tikzpicture}
\begin{axis}[
  width=\columnwidth, height=4.6cm,
  scale only axis=false,
  ymin=0, ymax=80, ytick={0,20,40,60,80},
  xmin=0.45, xmax=4.55,
  xtick={1,2,3,4},
  xticklabels={{\deconlysr{}\\{\tiny Ettin 1B}}, {\dmrst{}\\{\tiny Ettin 1B}}, {\iclc{}\\{\tiny Qwen3.5 397B}}, {\iclf{}\\{\tiny DeepSeek V4 Pro}}},
  xticklabel style={align=center, font=\scriptsize},
  yticklabel style={font=\scriptsize},
  ylabel={Attachment accuracy (\%)}, ylabel style={font=\scriptsize},
  ymajorgrids, grid style={black!12},
  axis line style={black!40}, tick style={black!40},
  major x tick style={draw=none},
  legend style={font=\scriptsize, draw=none, fill=none,
                at={(0.5,1.04)}, anchor=south, legend columns=4, column sep=4pt},
  legend cell align=left,
  clip=false,
]
\addlegendimage{area legend, fill=black!55, draw=black!70}\addlegendentry{Explicit}
\addlegendimage{area legend, fill=black!12, draw=black!70, postaction={pattern=north east lines, pattern color=black!60}}\addlegendentry{Implicit}
\addlegendimage{black, thick, dashed}\addlegendentry{Written}
\addlegendimage{black, thick, densely dotted}\addlegendentry{Spoken}
\def\attachgroup#1#2#3#4#5#6#7#8#9{%
  \fill[#2] (axis cs:#1-0.36,0) rectangle (axis cs:#1-0.01,#3);
  \draw[#2!70!black, line width=0.4pt] (axis cs:#1-0.36,0) rectangle (axis cs:#1-0.01,#3);
  \fill[#2!22] (axis cs:#1+0.01,0) rectangle (axis cs:#1+0.36,#6);
  \fill[pattern=north east lines, pattern color=#2!80!black] (axis cs:#1+0.01,0) rectangle (axis cs:#1+0.36,#6);
  \draw[#2!70!black, line width=0.4pt] (axis cs:#1+0.01,0) rectangle (axis cs:#1+0.36,#6);
  \draw[black!80, line width=0.5pt] (axis cs:#1-0.185,#4) -- (axis cs:#1-0.185,#5)
        (axis cs:#1-0.235,#4) -- (axis cs:#1-0.135,#4) (axis cs:#1-0.235,#5) -- (axis cs:#1-0.135,#5);
  \draw[black!80, line width=0.5pt] (axis cs:#1+0.185,#7) -- (axis cs:#1+0.185,#8)
        (axis cs:#1+0.135,#7) -- (axis cs:#1+0.235,#7) (axis cs:#1+0.135,#8) -- (axis cs:#1+0.235,#8);
  \draw[black, thick, dashed] (axis cs:#1-0.42,#9) -- (axis cs:#1+0.42,#9);
}
\def\attachspoken#1#2{\draw[black, thick, densely dotted] (axis cs:#1-0.42,#2) -- (axis cs:#1+0.42,#2);}
\attachgroup{1}{decblue}{58.1}{55.0}{61.4}{48.1}{46.4}{50.0}{51.3}\attachspoken{1}{49.1}
\attachgroup{2}{dmrstorange}{72.4}{69.5}{75.5}{62.9}{61.2}{64.8}{66.1}\attachspoken{2}{63.5}
\attachgroup{3}{iclcviolet}{42.5}{39.4}{45.8}{38.9}{37.0}{40.8}{41.3}\attachspoken{3}{37.5}
\attachgroup{4}{iclfaqua}{62.7}{58.4}{67.0}{53.3}{50.8}{55.9}{55.3}\attachspoken{4}{56.3}
\end{axis}
\end{tikzpicture}
\caption{Attachment accuracy on GUM with gold EDUs, by relation explicitness (bars, 95\% CIs) and modality (lines). \iclf{} covers the 17 documents it parsed.}
\label{fig:gum-attach}
\end{figure}

\paragraph{Attachment across data types}
GUM marks whether each relation is signaled by a discourse marker (explicit) following the eRST framework \citep{zeldes-etal-2025-erst}, and covers spoken and written genres, so we compare attachment accuracy, i.e.~whether each EDU's dependency head is correct, along both dimensions (Figure~\ref{fig:gum-attach}).
Explicit relations are easier for every system, significantly so for all but \iclc{}, and written text is easier for all but \iclf{}.
In the matched Ettin pair, the encoder leads by about 15 points on explicit and implicit relations alike, and almost equally on spoken and written data, so its advantage does not come from relying on discourse markers.
The \iclf{} bars cover only the 17 of 32 documents it parsed without failure, which flatters it relative to the other systems.

\paragraph{Segmentation} The segmentation errors that separate the families fall at sentence-internal boundaries (Section~\ref{sec:results}).
Only GUM contains spoken data, and there speech is slightly harder to segment for most systems (e.g.\ $94.5$ vs.\ $95.0$ for \dmrst{} Ettin-1B, and $95.4$ vs.\ $96.2$ for \deconlysr{} Qwen~3.6 27B).
On GUM, errors made only by \dmrst{} with Ettin-1B, and not by \deconlysr{} with Qwen~3.6 27B, are mostly \textcolor{blue}{recall errors} (in \textcolor{blue}{blue}), sometimes involving complex elliptical environments, as in \ref{ex:gap-seg}, where only Qwen catches the blue segmentation point, or in \ref{ex:so}, which also has an Ettin \textcolor{red}{precision error} (marked in \textcolor{red}{red}, where this `so' actually belongs with `conceded so' in the same unit).

\ex. Speakers of Eegimaa use .. Gújjolaay .. || to refer to their language.., || but Eegimaa \textcolor{blue}{||} to distinguish their Jóola variety\label{ex:gap-seg}

\ex. the State, || and it can be found in the transcript .., \textcolor{blue}{||} admitted as much and conceded \textcolor{red}{||} so in its closing argument \label{ex:so}

Outside of English, we note the poorest segmentation quality in Chinese, where XLM-R base struggles to decide when to separate fronted temporal adverbials, as in the false positive \ref{ex:zho-yesterday}. The somewhat better Qwen model also over-predicts boundaries around non-clausal phrases, as in \ref{ex:zho-death}.

\ex. \begin{CJK*}{UTF8}{gbsn}
昨天 \quad \textcolor{red}{||} \quad 在 \quad\quad  教育部 \quad \quad 举行 \quad  的 \\
yesterday \textcolor{red}{||} at education-ministry hold REL \\
``held at the education ministry \textcolor{red}{||} yesterday''\label{ex:zho-yesterday}
\end{CJK*}

\ex. \begin{CJK*}{UTF8}{gbsn}
在\textcolor{red}{||}逝世的\textcolor{red}{||} 第四天，捷克为他 举行 国葬 \\
at \textcolor{red}{||} death POSS \textcolor{red}{||} 4th day , Czechia for him held state funeral \\
``Czechia held a state funeral for him 4 days after his death''\label{ex:zho-death}
\end{CJK*}

\section{Discussion and conclusion}
\label{sec:conclusion}

Our results suggest several important takeaways. 
First, encoders should be the architecture of choice for gold EDU RST parsing: they outperform decoders at matched parameter budgets and match much larger models, a finding mirrored by recent results on other tasks, such as the latest shared task on coreference resolution, where an encoder-based system outperformed every LLM entry \citep{novak-etal-2025-findings,straka-2025-corpipe}.
Their advantage is clearer still on the much smaller non-English treebanks, where \dmrst{} wins in three of four languages, although differences in pretraining language coverage between XLM-R and the generative PLMs may also contribute. 

At the same time, large generative \textsc{sr} parsers show an edge in EDU segmentation in end-to-end settings, which are the realistic setting for discourse parsing.
On RST-DT, this edge widens the lead of models with around 30B parameters over our 1B encoder.
However, our apples-to-apples comparison using only Ettin models shows that even when used for segmentation, a pure encoder model outperforms a pure decoder at matched parameters.
Additionally, we observe that ICL approaches, even with near-frontier LLMs on the scale of DeepSeek V4 Pro (1.6T), cannot even beat an encoder with 150M parameters. 

These findings suggest that a golden middle path for practical RST parsing at this point may be to implement a decoder-based segmenter as an input to an encoder-based parser, or to develop hybrid models with shared parameters combining the strengths of these architectures.
That said, as we look at the gap between English results and lower-resourced languages, it remains likely that developing more data for supervision, especially for those languages, possibly coupled with stronger encoders beyond 1B should they become available, may bring the greatest gains.

Beyond providing these results, we release our most successful models alongside an easy-to-use software package for running them for the community, with the hope of enabling practical work using discourse parsing under realistic compute budgets, especially using the near-SotA Ettin encoder models reported in this paper. 

\section*{Limitations}

We do not perform an extensive hyperparameter search, given its computational cost: settings follow prevailing practice, and tuning within each family was restricted to RST-DT, confined to development set performance, and used only a single PLM, so further tuning may in some cases yield better results.
All results come from a single training run per configuration, and run-to-run variance is non-trivial, especially on small datasets: for example, runs of the German discriminative parser differ by up to $3.5$ points of gold-EDU Full.
Our English corpora are several times larger and should be less at risk, but we have not carefully measured their variance.

We also did not test for data contamination.
GUM, including its gold RST trees, is publicly available, and RST-DT trees circulate in public parser repositories, so any of our PLMs may have seen test annotations during pretraining.

\bibliography{custom}

\appendix

\section{Data preprocessing}\label{app:preproc}

In RST-DT, one test document, \texttt{wsj\_1189}, carries an irregularity in the released annotation.
We resolve it to the reading its LDC source implies and keep the document, evaluating on the full 38-document test set.

Each of the four non-English datasets required some interventions, which were necessary and which we record in full since these alter the data a system is trained or scored on. Where possible, we oriented our decisions to ones taken in the most recent edition of the shared task on Discourse Relation Parsing and Treebanking (DISRPT 2025, \citealt{braud-etal-2025-disrpt}), which contains all of the datasets used in this paper, but does not provide full discourse trees. 
Instead, the task's relation classification track lists pairs of pre-segmented text spans and challenges participants to predict the relations between them.

For German, every PCC document is annotated as a two-root forest in which the headline is left unattached. 
DISRPT resolves this by dropping the headline, and we follow. 
One further document is disconnected even after that, which we exclude, leaving us with 141 of the original 142 training documents to use.

For Persian, one document is likewise annotated as two adjacent top-level spans that were never joined, and because it falls in the test split, we join them under a multinuclear \textsc{joint} rather than discard it.

In Basque, one test document contains a duplicated node whose two copies are identical.
We deduplicate these nodes.
Further, one development document is left almost entirely unannotated---only four of its sixteen segments attach to anything at all, and DISRPT's own release lists a single relation for it.
We also drop this document, yielding 23 of the original 24 development documents.

Finally, Basque and German each contain multinuclear nodes carrying more than one satellite.
We read these in the standard nested way, interposing a span node so that the multinuclear core takes its nearest satellite, the resulting span takes the next, and so on, which reduces the structure to the single-satellite case these corpora otherwise use throughout. 
This affects $21$ multinuclear nodes and interposes $23$ spans, and because no multinuclear node in either English corpus carries more than one satellite it cannot alter our English trees.

Two smaller repairs are confined to training data: five Persian documents carry a seemingly meaningless multinuclear node with a single child, which we collapse into its grandparent, and one Basque document types as a span a node whose two children are both satellites, which we read as multinuclear.

Chinese GCDT needs no structural repair, and we take its text as distributed, in word-segmented form, as is standard for this corpus.
This setting is, however, somewhat unrealistic, since natural Chinese has no spaces between words, and e2e algorithms would potentially need to predict EDU boundaries at the level of individual Chinese characters.
GCDT's trees do escape square brackets in the PTB style as \texttt{-LSB-} and \texttt{-RSB-}, which we restore to the literal characters, as in DISRPT, occurring in $26$ of its $50$ documents.

\section{Discriminative parsers}
\label{app:disc}

\subsection{Encoder choice}
\label{app:enc-size}
While encoders larger than 1B do exist, evidence suggests that they would not offer better representations here.
XLM-R XL (3.5B) and XXL (10.7B) are multilingual instead of monolingual, and the large Ettin models score higher on MNLI and match or exceed them on SST-2, the two benchmarks shared between the two suites' evaluations \citep{goyal-etal-2021-larger,weller2025seqvsseqopen}.
EuroBERT \citep{eurobert} 2.1B is likewise multilingual, and on the English portions of its own evaluation, including NLI and NER, it performs only comparably to five-times-smaller ModernBERT-large \citep{warner-etal-2025-smarter}, the English-only model whose recipe Ettin replicates and slightly exceeds at matched size.
We choose XLM-R base rather than large because large diverged in two of five German training runs.

\subsection{Recipe}
\label{app:disc-recipe}
The exact settings of the RST-DT runs are given below. 
Each system's GUM run shares this recipe, differing only in the training data and the fine-grained relation inventory.
The discriminative parsers train under bf16 autocast with fp32 weights and evaluate in fp32.
Table~\ref{app:tab:disc} lists hyperparameters for main experiments.
The encoder is adapted with LoRA, with validation every epoch.
The learning rate schedule is linear warmup followed by linear decay to zero, identical to \texttt{transformers}' \texttt{get\_linear\_schedule\_with\_warmup}.
As is standard, documents are encoded whole, with a ``batch'' actually consisting of a single whole document, with gradients accumulated as shown.

\paragraph{Head architectures}
The biaffine heads of \topdown{} and \shiftreduce{} have 512 hidden units.
\shiftreduce{} also chooses each action with a 512-unit feedforward network over the top two discourse units on the stack and the next EDU in the queue.
\dmrst pairs a single-layer GRU decoder and dot-product pointer attention with BIE/CRF segmentation and dynamic loss weighting (temperature 2, window 30).

\paragraph{Model selection and early stopping}
The checkpoint with highest Full on the development set is used for final evaluation.
Training stops early if this metric does not improve for 10 epochs.

\begin{table}[t]
\centering
\small
\begin{tabular}{lccc}
\toprule
 & \topdown & \shiftreduce & \dmrst \\
\midrule
Optimizer & \multicolumn{3}{c}{AdamW} \\
Weight decay & \multicolumn{3}{c}{0.01} \\
LoRA rank  & \multicolumn{3}{c}{16} \\
LoRA $\alpha$ &  \multicolumn{3}{c}{32} \\
LoRA dropout & \multicolumn{3}{c}{0.05} \\
LoRA targets & \multicolumn{3}{c}{all linear} \\
Adapter LR & \multicolumn{3}{c}{$3{\times}10^{-4}$} \\
Head LR & \multicolumn{3}{c}{$2{\times}10^{-4}$} \\
Warmup steps & 1000 & 1000 & 100 \\
Grad.\ clip & 1.0 & 1.0 & 20.0 \\
Grad.\ accum. & 1 & 1 & 3 \\
Epochs & 50 & 50 & 100 \\
Patience & \multicolumn{3}{c}{10} \\
Dropout & 0.2 & 0.2 & 0.5$^{\dagger}$ \\
\bottomrule
\end{tabular}
\caption{Discriminative parser settings for the main table (LoRA) runs.
No LoRA bias terms are trained and DoRA is off.
$^{\dagger}$\dmrst dropout: 0.5 (encoder, decoder, labeler), 0.2 (document GRU), 0.3 (segmenter).}
\label{app:tab:disc}
\end{table}

\subsection{Full fine-tuning vs.\ LoRA}
\label{app:disc-repro}

The main table discriminative results use LoRA so that adaptation is held fixed across the PLM sizes and matched to the generative family.
This is a departure from prior work, which typically fully fine-tunes the encoder at a low ($\text{<}1\text{e-}3$) learning rate.

To confirm that discriminative parsers fine-tuned with LoRA remain competitive with full fine-tuning, we compare the two strategies for \topdown{} and \shiftreduce{} with XLM-R large (550M) and for \dmrst{} with XLM-R base (270M) and large.
The full fine-tuning runs use encoder learning rates of $1 \times 10^{-5}$ (\topdown, \shiftreduce) and $2{\times}10^{-5}$ (\dmrst), and otherwise match the recipe shown in Table \ref{app:tab:disc}.
Overall, both LoRA and full fine-tuning lead to good results on RST-DT, with the largest difference being 0.8 absolute Full points.

\begin{table}[t]
\centering
\small
\begin{tabular}{lcccc}
\toprule
 & S & N & R & F \\
\midrule
\multicolumn{5}{l}{\topdown{} (XLM-R large)} \\
\quad FFT & 78.6 & 66.5 & 56.5 & 54.3 \\
\quad LoRA & 77.4 & 66.2 & 56.3 & 54.1 \\
\addlinespace
\multicolumn{5}{l}{\shiftreduce{} (XLM-R large)} \\
\quad FFT & 76.5 & 66.2 & 55.2 & 53.5 \\
\quad LoRA & 76.9 & 66.6 & 56.4 & 54.2 \\
\addlinespace
\multicolumn{5}{l}{\dmrst{} (XLM-R base)} \\
\quad FFT & 75.7 & 64.6 & 54.0 & 52.6 \\
\quad LoRA & 74.7 & 64.6 & 53.9 & 52.6 \\
\addlinespace
\multicolumn{5}{l}{\dmrst{} (XLM-R large)} \\
\quad FFT & 78.1 & 67.6 & 56.9 & 55.7 \\
\quad LoRA & 76.6 & 66.9 & 56.4 & 54.9 \\
\bottomrule
\end{tabular}
\caption{Discriminative parsers under full fine-tuning (FFT) and under the LoRA recipe of Table~\ref{app:tab:disc}, on RST-DT, gold EDU.}
\label{app:tab:disc-repro}
\end{table}

\section{Generative parsers}
\label{app:gen}

\subsection{Recipe}
\label{app:gen-recipe}
The exact settings of the RST-DT runs are given below, and each system's GUM run shares this recipe, differing only in the training data and the fine-grained relation inventory.
The generative parsers hold their frozen base in bf16 and both train and decode at that precision.
Table~\ref{app:tab:gen-shared} lists the settings shared by all generative parsers.
The exact PLMs are \texttt{google/t5gemma-2-1b-1b} (\encdecsr, \encdecsexp), \texttt{google/gemma-3-1b-it}, \texttt{google/gemma-4-31B-it}, and \texttt{Qwen/Qwen3.6-27B} (\deconlysr, \deconlysexp).

We use the post-trained release of each PLM and present the source as a raw text stream, ignoring its chat template.
We use Adafactor rather than AdamW for its much smaller memory footprint, without which models above 10B parameters would have been prohibitively memory-intensive to train on our hardware.
There is only one exception to the values shown in the table: the Qwen 3.6 27B S-expression run has a higher maximum output length of 24576, which was necessary in order to ensure untruncated sequences.

\paragraph{Trained parameters}
The LoRA adapters are the only trained PLM weights.
Each model's vocabulary is extended with additional tokens---the serialization alphabets described in Appendix~\ref{app:serial}, plus, for decoder-only PLMs, a separator token marking the boundary between the document and the action stream.
These new vocabulary embeddings are trained, and all other embeddings remain frozen.

\paragraph{Model selection and early stopping}
The checkpoint with highest Full on the development set is used for final evaluation.
Validation cycles are very expensive due to the autoregressive nature of these models, and we therefore perform them only every other epoch of the last stage of the training curriculum.
Training stops early if the validation metric does not improve after 5 validation cycles (i.e.\ 10 epochs).

\begin{table}[t]
\centering
\small
\begin{tabular}{ll}
\toprule
Setting & Value \\
\midrule
Optimizer & Adafactor$^{\ddagger}$ \\
Learning rate & $3\times10^{-4}$ \\
LR schedule & warmup--stable--decay (see text) \\
Weight decay & 0.05 \\
Label smoothing & 0.1 \\
Gradient clip & 1.0 \\
Effective batch & 8 docs ($1{\times}8$ accum.) \\
Grad.\ checkpointing & yes \\
Max input length & 16384 \\
Max output length & 16384 \\
LoRA rank & 16 \\
LoRA $\alpha$ & 32 \\
LoRA dropout & 0.1 \\
LoRA targets & all linear modules$^{\dagger}$ \\
Patience & 5 validations (10 epochs) \\
\bottomrule
\end{tabular}
\caption{Hyperparameters shared by all generative parsers.
$^{\ddagger}$Adafactor with \texttt{relative\_step} and \texttt{scale\_parameter} disabled, so the external schedule sets the rate.
$^{\dagger}$On the multimodal Gemma 4 31B, the adapter regex targets the text stack only, excluding the vision and audio towers.}
\label{app:tab:gen-shared}
\end{table}

\paragraph{Curriculum}
All generative models use the same curriculum, tuned in preliminary experiments.
We adopt this regimen because of instabilities and divergences we observed in generative parsers when full documents were given from the beginning, especially under S-expression serialization.
This curriculum consists of a five-phase schedule of 60 epochs, with EDU caps of 8, 16, 32, and 64 for 8, 4, 4, and 4 epochs, then full documents for 40.
Each subtree stage trains on the maximal subtrees within its cap, subsampled uniformly to at most 2 times the document count.
The learning rate schedule is customized to this curriculum: there is a linear warmup over the first 200 steps, then the learning rate holds at maximum across all stages before the last, and then in the final stage, another linear warmup of 150 steps is applied, followed by a linear decay to 10\% of peak learning rate over the remaining epochs.

\subsection{Serialization and constrained decoding}
\label{app:serial}
We illustrate throughout with the three-EDU document of Figure~\ref{fig:serial}.
Both serializations reproduce the source text through a \texttt{<copy>} action, which copies the token at a cursor and advances the cursor by one position.
For both serialization schemes, all possible decodes are syntactically valid by construction under our constrained decoding approach.

\subsubsection{Shift-reduce}
\label{app:serial-sr}

The output alphabet is a small action set: \texttt{<copy>}, \texttt{<shift>}, and one fused \texttt{<reduce\_*>} token per nuclearity--relation pair, scored by a replacement head over that alphabet.
This head replaces the pretrained language modeling head, and each of its rows is warm-initialized from the corresponding unembedding row.
No terminator or trie is needed, since a reduce with its labels is a single action token.

The decode state is the source cursor, the stack size, and the length of the currently open EDU.
\texttt{<copy>} is legal while source remains, \texttt{<shift>} once the open EDU is non-empty, and \texttt{<reduce>} once the stack holds two subtrees.
\texttt{<eos>} is legal only once all the source tokens have been consumed, with exactly one item on the stack.

Under the gold EDU condition, the model is forced to copy up to each gold boundary and \texttt{<shift>} at the boundary.

\subsubsection{S-expression}
\label{app:serial-sexp}

\textsc{sexp} resembles the scheme of \citet{hu-wan-2023}, but places each node's nuclearity--relation label after its children rather than before them, and uses curly rather than square brackets.
Following their hypothesis that labels spelled out in words are easier to learn given pretraining than new special tokens, we render labels as ordinary words.

The output alphabet is the LM's own subword vocabulary together with \texttt{<copy>} and a \texttt{<label\_end>} terminator, scored by the LM's own pretrained head.
Each internal node's label is spelled out as its nuclearity code and relation in words, followed by \texttt{<label\_end>}, as in \texttt{SN elaboration <label\_end>}.
The terminator marks where a label ends, and is necessary because some relation names are prefixes of others, such as with \texttt{elaboration} and \texttt{elaboration additional}.
Wherever a label is required, the decoding mask follows a trie over the relation inventory, admitting only tokens that continue some label until a complete label is recognized.

\paragraph{Legality}
Several invariants guide the constrained decoding applied for S-expression validity, which we detail here.
Every open bracket becomes either a leaf, which contains copied text, or an internal node, which contains two children and a label.
The first action inside a new bracket decides which: a \texttt{<copy>} makes it a leaf, and another open bracket makes it internal.
A leaf may close once it contains at least one subword.
An internal node may close once it has two children and a label.
The root may close only once the entire source has been copied.

Throughout decoding, the number of leaves still required may never exceed the number of source subwords remaining, since each leaf needs at least one to close without leaving an EDU empty.
This rules out opening a bracket that would require too many leaves (with one subword left, the model may open a leaf but not an internal node, which needs two) and copying a subword that a required leaf still needs.
Likewise, the last required leaf cannot close while source remains, so every subword ends up in exactly one leaf.

\paragraph{Gold EDU forcing}
Under the gold EDU condition, decoding is further constrained so that every leaf is exactly one gold EDU.
Copying within a leaf is forced up to the gold boundary, and each node carries a budget of gold EDUs to cover: the root covers the whole document, a split divides a node's budget between its children, and a budget of one forces a leaf while a larger one forces an internal node.

Unlike \textsc{sr}, which reduces only over completed EDUs, \textsc{sexp} performs segmentation and structure prediction at the same time.
This can lead to issues if the model is forced into a segmentation that it did not anticipate: for example, if the model has opened brackets for a constituent of one EDU but gold forcing inserts a boundary within it, the extra EDU must be fitted into whatever brackets are still open, and the tree around it is no longer the one the model was trying to build.
Forcing can therefore have more deleterious effects on \textsc{sexp} models compared to \textsc{sr} models.

\subsection{Beam search}
\label{app:beam}

\begin{table*}[t]
\centering
\small
\begin{tabular}{llccccc}
\toprule
& & \multicolumn{2}{c}{Seg} & \multicolumn{2}{c}{F} & \\
\cmidrule(lr){3-4}\cmidrule(lr){5-6}
\textbf{System} & \textbf{PLM} & greedy & beam & greedy & beam & $\Delta$F \\
\midrule
\encdecsr & T5Gemma 2 1B-1B & 97.8 & 97.9 & 51.4 & 52.2 & $+0.8$ \\
\encdecsexp & T5Gemma 2 1B-1B & 95.1 & 96.0 & 46.0 & 47.5 & $+1.5$ \\
\deconlysr & Gemma 3 1B & 96.7 & 96.6 & 45.7 & 45.2 & $-0.5$ \\
\deconlysexp & Gemma 3 1B & 92.5 & 94.0 & 34.4 & 34.3 & $-0.1$ \\
\deconlysr & Gemma 4 31B & 98.2 & 98.2 & 55.1 & 54.8 & $-0.4$ \\
\deconlysr & Qwen 3.6 27B & 98.2 & 98.2 & 54.9 & 54.9 & $-0.1$ \\
\deconlysexp & Gemma 4 31B & 94.9 & 96.8 & 47.5 & 49.5 & $+2.0$ \\
\deconlysexp & Qwen 3.6 27B & 95.8 & 96.4 & 47.5 & 48.3 & $+0.8$ \\
\bottomrule
\end{tabular}
\caption{Beam search ablation within the generative family: RST-DT end-to-end, identical checkpoints decoded greedily and with legality-constrained beam search (beam size 6).
Greedy is the setting used in Tables~\ref{tab:e2e} and~\ref{tab:goldedu}.
Segmentation and Full are shown per decoding, with $\Delta$F the beam-minus-greedy difference computed before rounding.}
\label{tab:beam}
\end{table*}

All systems decode greedily in main experiments in order to simply control for decoding budget across systems.
In Table~\ref{tab:beam}, we compare greedy and beam search (beam size 6) for our generative parsers.
Overall, we find that beam search provides modest benefits at best: at most 2 absolute points on segmentation and on Full.
In particular, our results show that shift-reduce systems appear to receive little benefit, and often slightly degrade under beam search, while S-expression systems always segment better, a gain which all but the weakest convert into modestly higher Full.

\section{In-context learning}
\label{app:icl}

\begin{figure*}[t]
\footnotesize
\hrule
\smallskip
\textbf{Task instructions, segmentation call}
\begin{verbatim}
You are doing Rhetorical Structure Theory (RST) discourse segmentation. Split the
document into Elementary Discourse Units (EDUs, roughly clause-level spans).

Reproduce the input document VERBATIM, inserting " || " at every boundary between
adjacent EDUs. Copy every word exactly, in order. Do NOT paraphrase, drop, merge,
or reorder any text. The ONLY characters you add are the " || " delimiters. EDUs
are fine-grained: split at clause boundaries, including subordinate clauses,
relative clauses, and many non-finite (to-infinitive and -ing) clauses, not just
at sentence boundaries.

Output ONLY the delimited text, no commentary.
\end{verbatim}
\textbf{Task instructions, parsing call}
\begin{verbatim}
You are doing RST discourse parsing. You are GIVEN the document's EDUs in order,
separated by " || ", and must produce ONLY the tree structure over them.
Nuclearity patterns:
  - NS: left child is the nucleus, right child is the satellite
  - SN: left child is the satellite, right child is the nucleus
  - NN: multinuclear, both children are nuclei (symmetric relations)

Emit a bottom-up SHIFT-REDUCE action sequence over the given EDUs, in order:
  - Emit the next EDU's words verbatim, then "<shift>", to push it (EDUs are
    consumed in the listed order, so the text you emit is fully determined).
  - A REDUCE combines the top two stack items into a parent, written as its
    nuclearity code (NS, SN, or NN), the relation spelled in words, then
    "<label_end>". For example "NS elaboration additional <label_end>".

For N EDUs there are exactly N "<shift>" actions and N-1 reduces. The only legal
reduce labels are:

- NS attribution <label_end>  (left child nucleus, right satellite)
- SN attribution <label_end>  (left child satellite, right nucleus)
- NS background <label_end>  (left child nucleus, right satellite)
[... one line per legal nuclearity-relation pair ...]

Output ONLY the action sequence, no commentary.
\end{verbatim}
\textbf{Prompt body, both calls}
\begin{verbatim}
Here are 5 worked examples.

--- Example 1 ---
INPUT: [example document]
OUTPUT: [its gold output]

[... examples 2-5 ...]

--- Now do this ---
INPUT: [the document to be processed]
OUTPUT:
\end{verbatim}
\hrule
\caption{The exact prompts used by both ICL systems.
A call's prompt is its task instruction block followed by the prompt body, whose INPUT and OUTPUT slots take the formats of Figure~\ref{fig:icl-sr}.
The parsing call's query INPUT is the segmentation call's output, or the gold segmentation in the gold EDU condition.
Bracketed text stands in for per-document material, the label inventory ($42$ lines for RST-DT, $57$ for GUM) is elided after its first three lines, and one over-length instruction line is wrapped for display.
Apart from the label inventory, the prompts are identical across corpora and both systems.}
\label{fig:icl-prompts}
\end{figure*}

Both ICL systems use the two-call pipeline of Section~\ref{sec:icl}: a segmentation call that reproduces the document with EDU delimiters, then a parsing call that emits a shift-reduce action sequence over the fixed EDU sequence.
Greedy decoding is used for all ICL experiments.
All reported rows use $k=5$ exemplars, selected after preliminary experiments on RST-DT development data.
Exemplars are drawn from the training split by one seeded permutation, of which the first $k$ are taken, so exemplar sets are nested across $k$ and a sweep measures $k$ rather than which documents were drawn.
The pool is restricted to documents of at most $60$ EDUs on RST-DT and $80$ on GUM so that the shared prompt prefix stays small.
No system prompt is used.
Figure~\ref{fig:icl-prompts} gives the exact prompts for both calls.

For \iclf{}, decoding is unconstrained.
We evaluate syntactically malformed outputs as completely incorrect for the given document.

\begin{figure}[t]
\centering
\small
\begin{tabular}{@{}p{0.96\columnwidth}@{}}
\toprule
$S_{0,0} \rightarrow$ \texttt{\footnotesize The cat sat . <shift>} $S_{1,1}$\\
$S_{1,1} \rightarrow$ \texttt{\footnotesize It was tired . <shift>} $S_{2,2}$\\
$S_{2,2} \rightarrow$ \texttt{\footnotesize So it slept . <shift>} $S_{3,3}$ $\mid$ \textsc{label} $S_{2,1}$\\
$S_{2,1} \rightarrow$ \texttt{\footnotesize So it slept . <shift>} $S_{3,2}$\\
$S_{3,3} \rightarrow$ \textsc{label} $S_{3,2}$\\
$S_{3,2} \rightarrow$ \textsc{label} $S_{3,1}$\\
$S_{3,1} \rightarrow \varepsilon$\\
\textsc{label} $\rightarrow$ \texttt{\footnotesize SN elaboration <label\_end>} $\mid$ \texttt{\footnotesize NS background <label\_end>} $\mid \cdots$\\
\bottomrule
\end{tabular}
\caption{The structure prediction grammar compiled for the three-EDU document of Figure~\ref{fig:icl-sr}.
Nonterminal $S_{s,d}$ is the parser state after $s$ shifts with $d$ subtrees on the stack, the start symbol is $S_{0,0}$, and derivations terminate at $S_{n,1}$.
Shift productions carry the text of the EDU they consume, and \textsc{label} expands to one string per nuclearity--relation pair.}
\label{fig:icl-grammar}
\end{figure}

\paragraph{Constrained decoding}
We devise a context-free grammar approach to constrain decoding for both stages of generative parsing (EDU segmentation and structure prediction).
Specifically, we devise a grammar specific to each document.
The grammar's nonterminals are parser states $S_{s,d}$, where $s$ and $d$ represent the number of shifts executed and the number of subtrees on the stack, respectively.
The accepting state $S_{n,1}$, where $n$ is the number of EDUs, emits the empty string, and all other states produce some combination of literal document tokens and shift and labeling actions.
In general, this context-free grammar approach is unambiguous and forces exactly $n$ shifts and $n-1$ reduces for $n$ EDUs.

An example is shown in Figure~\ref{fig:icl-grammar}.
We may observe that the first two states $S_{0,0}$ and $S_{1,1}$ force the parser to perform two shifts, and the first point at which the grammar offers a choice between actions is $S_{2,2}$, where the model may either reduce by emitting a \textsc{label} or continue shifting the third EDU.

We choose to make a shift's production include the text of the EDU it consumes under the hypothesis that this will allow the model to more effectively process the current state of decoding at any given point.
In a preliminary comparison with Qwen3.5-122B-A10B on gold EDUs, including EDU text like this instead of only decoding actions raised Full from $14.7$ to $28.1$ on the 29 RST-DT test documents that both variants parsed.

\paragraph{Context and reasoning}
For both \iclc{} models, we allow the model to perform unconstrained reasoning for up to $49{,}152$ tokens before reasoning is interrupted and the generation of the response under the grammar constraint is forced.
We use the full $262{,}144$-token context length for both models, except in the grammar ablation below, which ran with $65{,}536$ tokens per request.
Under this configuration, no document ever exceeds available context length during inference.

\paragraph{Grammar ablation}
In order to measure the effect of constrained decoding, we repeat our \iclc{} gold EDU experiments for the larger Qwen3.5-397B-A17B model but with the output grammar disabled.
This results in failures on $13$ of $38$ RST-DT documents and $30$ of $32$ GUM documents, causing Full to fall from $28.1$ to $12.4$ on RST-DT and from $20.1$ to $1.4$ on GUM.

\paragraph{Relation confusions}
\begin{figure*}[h!tb]
\centering
\includegraphics[width=\textwidth]{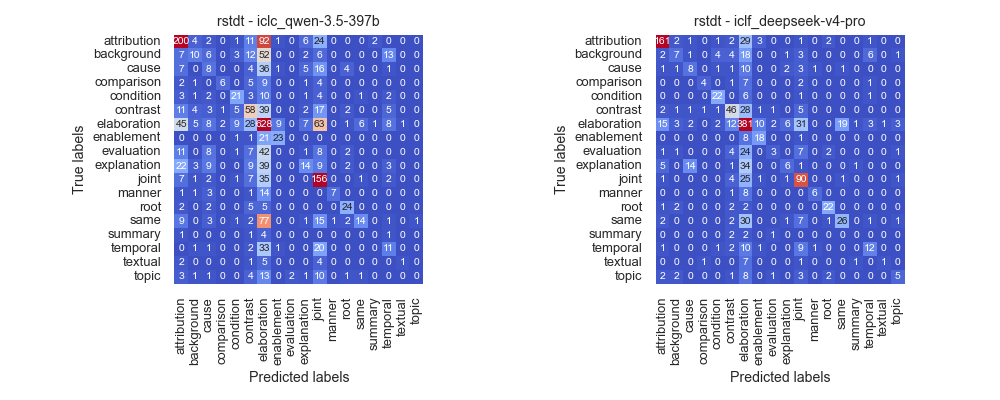}
\caption{Label confusion matrices for top ICL models on RST-DT.}
\label{fig:rstdt-icl-confmats}
\end{figure*}

Figure \ref{fig:rstdt-icl-confmats} shows the much less impressive performance of ICL approaches on discourse relation recognition for RST-DT.

\section{Segmentation and structure analysis}
\label{app:analysis}

Table~\ref{tab:seg-context} splits segmentation recall by what precedes each gold boundary.
Boundaries after sentence-final punctuation are recovered almost perfectly by every system, so all differences lie at sentence-internal boundaries.
On RST-DT the large pretrained PLMs lead most where no punctuation cues the boundary, as in complement and attribution clauses.
On GUM all differences are small, and the large generative parsers gain on both recall and precision.

Table~\ref{tab:width} gives gold EDU span recall by constituent width, excluding each document's root.
All parsers are close on two-EDU constituents, and the generative parsers fall behind as width grows.
With relation and nuclearity also required, the deficit on wide GUM constituents is larger still: at 17--32 EDUs \dmrst{} recovers $16.0$--$18.6\%$ of constituents fully, the 27--31B parsers $7.4$--$11.2\%$, and the Gemma~3 and T5Gemma parsers $4.3$--$5.3\%$.

\begin{table}[t]
\centering
\footnotesize
\setlength{\tabcolsep}{2.5pt}
\begin{tabular}{lcccccc}
\toprule
& \multicolumn{3}{c}{RST-DT} & \multicolumn{3}{c}{GUM} \\
\cmidrule(lr){2-4}\cmidrule(lr){5-7}
& sent. & punct. & none & sent. & punct. & none \\
\midrule
\dmrst{} 400M & 99.9 & 95.9 & 93.4 & 99.8 & 94.2 & 92.6 \\
\dmrst{} 1B & 99.8 & 97.7 & 92.8 & 99.9 & 92.7 & 91.4 \\
Ettin-1B dec. & 99.7 & 96.8 & 91.0 & 99.9 & 89.6 & 86.5 \\
Gemma 3 1B & 99.7 & 97.3 & 93.6 & 99.7 & 91.3 & 89.6 \\
T5Gemma 2 & 99.8 & 98.8 & 96.7 & 99.9 & 92.8 & 92.6 \\
Gemma 4 31B & 99.9 & 98.8 & 97.9 & 99.9 & 93.7 & 92.7 \\
Qwen 3.6 27B & 99.9 & 98.8 & 97.0 & 100.0 & 94.2 & 93.1 \\
\midrule
$n$ & 924 & 560 & 824 & 1368 & 1028 & 1351 \\
\bottomrule
\end{tabular}
\caption{End-to-end segmentation recall by the token preceding each gold boundary: sentence-final punctuation (sent.), other punctuation (punct.), or none.
Predictions are projected onto gold tokens as in Section~\ref{sec:eval}, the generative rows are \textsc{sr} parsers, the Ettin-1B decoder (dec.) is the \deconlysr{} parser of Table~\ref{tab:arch}, and $n$ is the number of gold boundaries.}
\label{tab:seg-context}
\end{table}

\begin{table*}[t]
\centering
\footnotesize
\setlength{\tabcolsep}{4pt}
\begin{tabular}{lcccccccccccc}
\toprule
& \multicolumn{6}{c}{RST-DT} & \multicolumn{6}{c}{GUM} \\
\cmidrule(lr){2-7}\cmidrule(lr){8-13}
Width (EDUs) & 2 & 3--4 & 5--8 & 9--16 & 17--32 & 33+ & 2 & 3--4 & 5--8 & 9--16 & 17--32 & 33+ \\
\midrule
\dmrst{} 400M & 92.7 & 85.8 & 70.7 & 52.9 & 47.6 & 29.1 & 89.6 & 80.6 & 71.4 & 61.8 & 52.7 & 46.6 \\
\dmrst{} 1B & 93.8 & 85.7 & 70.7 & 58.7 & 54.5 & 33.0 & 90.0 & 80.6 & 71.7 & 58.5 & 44.7 & 44.7 \\
Ettin-1B dec. & 91.9 & 76.7 & 46.7 & 24.0 & 16.1 & 32.0 & 84.0 & 68.5 & 41.1 & 26.7 & 15.4 & 28.1 \\
Gemma 3 1B & 91.3 & 78.8 & 52.8 & 32.7 & 24.5 & 33.0 & 84.4 & 71.4 & 46.2 & 22.8 & 16.0 & 32.8 \\
T5Gemma 2 & 93.2 & 84.9 & 61.2 & 42.3 & 37.8 & 26.2 & 90.4 & 77.9 & 58.3 & 36.2 & 23.4 & 37.2 \\
Gemma 4 31B & 93.8 & 88.1 & 69.1 & 50.5 & 33.6 & 31.1 & 90.0 & 81.5 & 63.7 & 42.3 & 30.9 & 42.7 \\
Qwen 3.6 27B & 93.2 & 86.0 & 72.8 & 53.8 & 44.8 & 25.2 & 90.9 & 83.0 & 68.2 & 51.3 & 47.3 & 30.0 \\
\midrule
$n$ & 780 & 657 & 379 & 208 & 143 & 103 & 1215 & 1000 & 669 & 390 & 188 & 253 \\
\bottomrule
\end{tabular}
\caption{Gold EDU span recall by constituent width in EDUs, excluding each document's root. Rows are as in Table~\ref{tab:seg-context}, and $n$ is the number of gold constituents.}
\label{tab:width}
\end{table*}

\section{Multilingual comparison with prior work}
\label{app:multiling-prior}

For the non-English treebanks, the closest point of comparison is the treebank-specific baselines of \citet{chistova-2025-bridging}, which use the same DISRPT 2025 partitions and original Parseval.
With gold EDUs, their Full scores are $23.6$ for German, $40.1$ for Persian, and $50.7$ for Chinese, against $23.1$, $37.1$, and $44.9$ for our \dmrst{} (Table~\ref{tab:multiling}).
The comparison is indicative rather than matched: their models use XLM-R large rather than base and average three seeds, and their label inventories are coarser, merging infrequent German and Persian relations following \citet{braud-etal-2017-cross} and keeping GCDT's coarse classes where we use its 32 fine ones.
Their Basque results come from the 88-document open release of the treebank rather than the DISRPT partition and are not comparable to ours.

\section{Compute, software, and artifacts}
\label{app:compute}

The discriminative and ${\leq}4$B generative models were each trained on a single NVIDIA A100 40GB GPU, and the 27--31B generative models on single H100 80GB or RTX PRO 6000 96GB GPUs.
The \iclc{} models were served with llama.cpp on two RTX PRO 6000 96GB GPUs, and \iclf{} was accessed through a commercial API gateway.
Including development and failed runs, the project used about 1,550 A100 GPU-hours and 570 H100 GPU-hours on the clusters, measured from their job accounting, and an estimated 600 further GPU-hours on the RTX PRO 6000 GPUs, which were not metered, about 100 of them for retraining the released models.

\paragraph{Software}
Experiments used Python 3.12 to 3.14, PyTorch 2.10 to 2.12, Hugging Face Transformers 5.9, and PEFT 0.19.
Tree metrics are computed by our own implementation and segmentation by the DISRPT 2025 scorer (Section~\ref{sec:eval}).

\paragraph{Licenses and intended use}
RST-DT is distributed by the Linguistic Data Consortium under its user agreement.
GUM's annotations are CC BY 4.0, and its texts carry the licenses of their sources, from CC BY to CC BY-NC-SA, with Reddit text available for non-commercial use only.
PCC is CC BY-NC-SA 4.0, and the GCDT repository is under Apache 2.0.
We found no license for the Persian corpus or for the Basque treebank.
We use all corpora for research only.
Ettin, XLM-R, and DeepSeek-V4 are MIT licensed, Qwen~3.5, Qwen~3.6, and Gemma~4 are Apache 2.0, and Gemma~3 and T5Gemma~2 are covered by the Gemma terms of use.
llama.cpp is MIT licensed.
We use all models within their licenses, for research.

\paragraph{Released artifacts}
Our code, under the MIT license, and ten trained models are released at \repo{}.
The released models were retrained on untokenized text so that they work on real text, so they are not the exact runs reported in our tables, and their model cards give their own scores.
They are intended for research and for analysis of text.
They carry the terms of their base models and of their training corpora, which the model cards state, so for example the German model is bound by the NonCommercial terms of PCC.

\section{AI Assistance Disclosure}
We used an AI coding assistant, Claude Code, to assist our implementation of these parsers to conduct experiments.
All code produced by the assistant was verified by the authors, and all experimental work conducted by the assistant was audited and verified by the authors.

\end{document}